\documentclass{article}
\usepackage{spconf}
\usepackage{hyperref}       
\usepackage{url}            
\usepackage{booktabs}       
\usepackage{amsfonts}       
\usepackage{nicefrac}       
\usepackage{microtype}      
\usepackage{xcolor}         
\usepackage{latexsym}
\usepackage{graphicx}
\usepackage{tikz}
\usepackage{amsthm,algorithmic}
\usepackage{epsfig,psfrag,amstext,amsfonts,amssymb,subeqnarray,color}
\usepackage{epstopdf,bm,multirow}
\usepackage{amsmath}
\usepackage{array,makecell,bm,threeparttable}  
\usepackage{marvosym}
\usepackage{ifsym}
\usepackage{subfig}
\usepackage{textcomp}
\usepackage{hyperref}
\usepackage{adjustbox}
\usepackage{nccfoots}
\usepackage[linesnumbered,ruled]{algorithm2e}
\usepackage{booktabs}
\usepackage{cuted}
\usepackage{flushend} 

\newtheorem{corollary}{Corollary}
\newtheorem{remark}{Remark}

\def\x{{\mathbf x}}
\def\y{{\mathbf y}}
\def\z{{\mathbf z}}
\def\f{{\mathbf{f}}}
\def\h{{\mathbf{h}}}
\def\u{{\mathbf u}}
\def\m{{\mathbf m}}

\def\vx{{\vec{\mathbf x}}}
\def\vy{{\vec{\mathbf y}}}
\def\vz{{\vec{\mathbf z}}}
\def\vf{{\vec{\mathbf f}}}
\def\vu{{\vec{\mathbf u}}}

\def\k{{\boldsymbol{K}}}

\def\cX{{\cal X}}

\def\cD{{\cal D}}

\def\cN{{\cal N}}

\def\bzeta{{\bm{\zeta}}}
\def\btheta{{{\bm{\theta}}}}

\definecolor{orange}{RGB}{200,0,100}

\title{Output-Dependent Gaussian Process State-Space Model
\vspace{-.1in}
}
\name{
Zhidi Lin\textsuperscript{$\dagger\ddagger$} \quad
Lei Cheng\textsuperscript{$\diamond$} \quad
Feng Yin\textsuperscript{$\dagger$}(\textrm{\Letter}) \quad
Lexi Xu\textsuperscript{$\circ$} \quad
Shuguang Cui\textsuperscript{$\dagger\ddagger$}
\thanks{
$^{\textrm{\Letter}}$ The corresponding author is Feng Yin (\textit{yinfeng@cuhk.edu.cn}). 
 }
\vspace{-.1in}
}
\address{
$\dagger$ 
School of Science and Engineering, The Chinese University of Hong Kong, Shenzhen, China
\\
$\ddagger$ Future Network of Intelligence Institute, The Chinese University of Hong Kong, Shenzhen, China
 \\
 $\diamond$  College of Information Science and Electronic Engineering, Zhejiang University, Hangzhou, China
 \\
 $\circ$ Research Institute, China United Network Communications Corporation, Beijing, China
   \vspace{-.1in}
}
\begin{document}
\topmargin=0mm
\ninept
\maketitle
\begin{abstract} 
\vspace{-.05in}
Gaussian process state-space model (GPSSM) is a fully probabilistic state-space model that has attracted much attention over the past decade. However, the outputs of the transition function in the existing GPSSMs are assumed to be independent, meaning that the GPSSMs cannot exploit the inductive biases between different outputs and lose certain model capacities. To address this issue, this paper proposes an output-dependent and more realistic GPSSM by utilizing the well-known, simple yet practical linear model of coregionalization (LMC) framework to represent the output dependency. To jointly learn the output-dependent GPSSM and infer the latent states, we propose a variational sparse GP-based learning method that only gently increases the computational complexity. Experiments on both synthetic and real datasets demonstrate the superiority of the output-dependent GPSSM in terms of learning and inference performance.
\end{abstract}
\vspace{-.03in}
%
\section{Introduction}
\label{sec:intro}
\vspace{-.1in}
A well-established probabilistic tool for modeling the underlying dynamical system of sequential data is the state-space model (SSM), which has been successfully applied in many areas of engineering, statistics, computer science, and economics \cite{sarkka2013bayesian}. For the case when the system dynamics 
are fairly known, a plethora of out-of-the-box learning and inference methods have been developed over the past decades, such as the Kalman filter (KF) for linear Gaussian dynamic systems, and the particle filter (PF) for nonlinear dynamic systems \cite{sarkka2013bayesian}. However, in some harsh scenarios, such as model-based reinforcement learning, or disease epidemic propagation, the underlying system dynamics cannot be well determined \textit{a priori} \cite{deisenroth2011pilco}.  Thus, the dynamics need to be learned from the observed noisy measurements, leading to the emergence of data-driven state-space models \cite{krishnan2017structured,alaa2019attentive,liu2022deep,mchutchon2015nonlinear, frigola2013bayesian, frigola2014variational,frigola2015bayesian,doerr2018probabilistic,ialongo2019overcoming}.

Gaussian processes (GPs), being an eminent Bayesian nonparametric models for machine learning  \cite{williams2006gaussian,suwandi2022gaussian,yin2020linear},  can be adopted as function priors in classical SSM, giving rise to the Gaussian process state-space model (GPSSM) \cite{ frigola2013bayesian}. A carefully selected GP prior provides not only meaningful uncertainty calibration in low data regime but also automatic scaling of model complexity based upon data volume \cite{theodoridis2020machine,cheng2022rethinking}. 
Due to these appealing properties, GPSSM and its variants have been applied to various applications, such as 
human motion capture
and pedestrian tracking and navigation \cite{wang2007gaussian,xie2020learning,yin2020fedloc,zhao2019cramer}.

Despite the ever-increasing popularity of GPSSM, accurate, simultaneous learning and inference in GPSSM remains a challenging problem. 
Much progress has been made over the past decade along different paths \cite{mchutchon2015nonlinear, frigola2013bayesian, frigola2014variational,frigola2015bayesian,doerr2018probabilistic,ialongo2019overcoming,lindinger2022laplace,eleftheriadis2017identification}.  Concretely, the first fully Bayesian learning of GPSSM was proposed in
\cite{frigola2013bayesian} using particle Markov chain Monte Carlo. 
Variational inference methods were developed based upon the mean-field (MF) assumption to reduce the heavy computational load of the sampling methods \cite{mchutchon2015nonlinear,frigola2014variational,eleftheriadis2017identification}.
More recent works have been devoted to overcoming the MF assumption for enhanced learning and inference performance \cite{doerr2018probabilistic,ialongo2019overcoming,lindinger2022laplace,liu2021gaussian}. However, all the existing methods utilize independent GPs to model the multi-outputs of the transition function while ignoring their dependencies, which can cause inductive bias between the outputs that cannot be transferred to improve the model generalization \cite{caruana1997multitask}.  
The inference performance can be significantly degraded, especially when the latent states are only partially observed (see Section \ref{subsec:syntheticdata}).  Moreover, high-dimensional data features nowadays are often entangled. For instance, in disease progression prediction application, the disease state of a patient comprises a series of mutually influencing physiological metrics \cite{alaa2019attentive}. Therefore, assuming the independence of outputs is simplifying but unrealistic.

In this paper, we aim to address the above-mentioned issues by explicitly modeling the output dependency without sacrificing much
computational complexity. The main contributions are summarized as follows.    
First, we resort to a simple yet practical framework, namely the linear model of coregionalization (LMC) \cite{alvarez2012kernels,chen2022multitask} to encode dependency among outputs of the GP-based transition function in GPSSM. To the best of our knowledge, this is the first study on output-dependent GPSSMs.
Second, we propose a variational learning method based upon the sparse GP \cite{hensman2013gaussian}, in which learning and inference only gently increase the computational complexity. 
Third, experimental results obtained using  real and synthetic datasets corroborate that the proposed output-dependent GPSSM outperforms various benchmark methods, including the output-independent GPSSM \cite{doerr2018probabilistic} and the deep state-space model (DSSM) \cite{krishnan2017structured}.

The remainder of this paper is organized as follows. Some preliminaries related to GPSSM are provided in Section \ref{sec:preliminaries}. In Section \ref{sec:proposed-model}, we introduce our proposed output-dependent GPSSM and detail the learning and inference algorithm. Numerical results are provided in Section \ref{sec:experimental-results}. Finally, we conclude the paper in Section \ref{sec:conclusion}.

\vspace{-.11in}
\section{Preliminaries}
\label{sec:preliminaries} 

\vspace{-.12in}
\subsection{Gaussian Process}
\label{subsec:GP-review} \vspace{-.1in}
Gaussian process (GP) defines a collection of random variables indexed by $\cX \subseteq \mathbb{R}^{d_x}$, such that any finite collection of these variables follows a joint Gaussian distribution \cite{williams2006gaussian}. With this definition, a Gaussian process is typically used to define a distribution over functions $f(\x): \mathbb{R}^{d_x} \mapsto \mathbb{R}$, 
\begin{equation}
\setlength{\abovedisplayskip}{3.5pt}
\setlength{\belowdisplayskip}{3.5pt}
    f(\x) \sim \mathcal{G} \mathcal{P}\left(\mu(\x), \  k_{\boldsymbol{\theta}_{gp}}\left(\x, \x^{\prime} \right)\right),
\end{equation}
where $\mu(\x)$ is a mean function, usually set to be zero in practice; $k_{\boldsymbol{\theta}_{gp}}\left(\x, \x^{\prime}\right)$ is a covariance function/kernel function; $\boldsymbol{\theta}_{gp}$ is a set of hyperparameters to be tuned for model selection. Following Bayes' theorem, the function prior is combined with new data to obtain an analytical posterior distribution. More specifically, given a noise-free training dataset $\cD = \{X, \f \}=\{\x_{i}, \mathrm{f}_i\}_{i=1}^n$, the posterior distribution $p(f(\x_*) \vert \x_*, \cD)$ at any test input $\x_* \in \cX$ is Gaussian, fully characterized by the posterior mean $\xi$ and the posterior variance $\Xi$:
\begin{subequations}
\setlength{\abovedisplayskip}{4.5pt}
\setlength{\belowdisplayskip}{4.5pt}
\begin{align}
 & \xi(\x_*) = \mu(\x_*) + \bm{K}_{{\x}_*, X} \bm{K}_{X,X}^{-1} \left(\f - \mu(X)\right), \label{eq:posterior_mean}\\
 & \Xi(\x_*) = k(\x_*, \x_*) - \bm{K}_{{\x}_*, X} \bm{K}_{X,X}^{-1} \bm{K}_{{\x}_*, X}^\top, \label{eq:posterior_vairiance}
\end{align}
\end{subequations}
where $\bm{K}_{X,X}$ denotes the covariance matrix evaluated on the training input ${X}$, and each entry is $[\bm{K}_{X,X}]_{i,j} = k_{\btheta_{gp}}({\x}_i, {\x}_j)$; $\bm{K}_{{\x}_*, X}$ denotes the cross covariance matrix between the test input ${\x}_*$ and the training input ${X}$;  $\mu(X) = \{\mu(\x_i)\}_{i=1}^n$ denotes the prior mean function evaluated on $X$.

\vspace{-.1in}
\subsection{Gaussian Process State-Space Model}
\label{subsec:GPSSM-review} \vspace{-.05in}
A generic state-space model (SSM) describes the probabilistic dependence between latent state $\x_t \in \mathbb{R}^{d_x}$ and observation $\y_t \in \mathbb{R}^{d_y}$.  Mathematically, it can be written as
\begin{subequations}
\setlength{\abovedisplayskip}{3.5pt}
\setlength{\belowdisplayskip}{3.5pt}
    \begin{align}
        & \x_{t+1} = f(\x_t) + \mathbf{v}_t,\\
        & \y_{t} = g(\x_t) + \mathbf{e}_t,
    \end{align}
\end{subequations} 
where 
$\mathbf{v}_t$ and $\mathbf{e}_t$ are additive noise terms; $f(\cdot)$
and $g(\cdot)$
are \textit{transition function} and \textit{emission function}, respectively.  
%
%

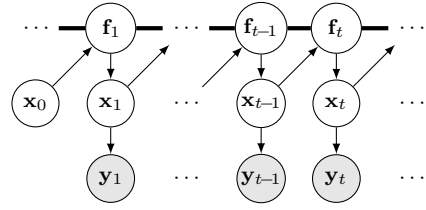
\begin{figure}[t!]
	\centering
	\footnotesize
	\begin{tikzpicture}[align = center, latent/.style={circle, draw, text width = 0.33cm}, observed/.style={circle, draw, fill=gray!20, text width = 0.33cm}, transparent/.style={circle, text width = 0.33cm}, node distance=1cm]
		\node[latent](x0) {${\x}_0$};
		\node[latent, right of=x0](x1) {${\x}_{1}$};
		\node[transparent, right of=x1](x2) {$\cdots$};
		\node[latent, right of=x2](xt-1) {$\!\!{\x}_{t\!-\!1}\!\!$};
		\node[latent, right of=xt-1](xt) {${\x}_{t}$};
		\node[transparent, right of=xt](xinf) {$\cdots$};
		\node[transparent, above of=x0](f0) {$\cdots$};
		\node[latent, above of=x1](f1) {${\f}_{1}$};
		\node[transparent, right of=f1](f2) {$\cdots$};
		\node[latent, above of=xt-1](ft-1) {$\!\!{\f}_{t\!-\!1}\!\!$};
		\node[latent, above of=xt](ft) {${\f}_{t}$};
		\node[transparent, right of=ft](finf) {$\cdots$};
		\node[observed, below of=x1](y1) {${\y}_{1}$};
		\node[transparent, below of=x2](y2) {$\cdots$};
		\node[observed, below of=xt-1](yt-1) {$\!\!{\y}_{t\!-\!1}\!\!$};
		\node[observed, right of=yt-1](yt) {${\y}_{t}$};
		\node[transparent, right of=yt](yinf) {$\cdots$};
		\draw[-latex] (x0) -- (f1);
		\draw[-latex] (f1) -- (x1);
		\draw[-latex] (x1) -- (f2);
		\draw[-latex] (ft-1) -- (xt-1);
		\draw[-latex] (xt-1) -- (ft);
		\draw[-latex] (x2) -- (ft-1);
		\draw[-latex] (ft) -- (xt);
		\draw[-latex] (xt) -- (finf);
		\draw[-latex] (x1) -- (y1);
		\draw[-latex] (xt-1) -- (yt-1);
		\draw[-latex] (xt) -- (yt);
		\draw[ultra thick]
		(f0) -- (f1)
		(f1) -- (f2)
		(f2) -- (ft-1)
		(ft-1) -- (ft)
		(ft) -- (finf)
		;
	\end{tikzpicture}
	\caption{Graphical model of GPSSM.
    The thick horizontal bar represents a set of fully connected nodes, i.e., the GP.}
    \vspace{-.25in}
    \label{fig:graphical_model}
\end{figure}

Placing a GP prior over the transition function $f(\cdot)$ and assuming a parametric emission function $g(\cdot)$ in SSM leads to the well-known Gaussian process state-space model (GPSSM)\footnote{The GPSSM considered in this paper keeps the same model capacity as the \textit{ones with both transition and emission GPs} while avoiding the severe \textit{non-identifiability} issue.
One can refer to \cite{frigola2015bayesian} (Section 3.2.1) for more details.} \cite{frigola2015bayesian}, which is depicted in Fig.~\ref{fig:graphical_model} and expressed mathematically as:
\begin{subequations}
\setlength{\abovedisplayskip}{3.5pt}
\setlength{\belowdisplayskip}{3.5pt}
    \label{eq:gpssm}
    \begin{align}
        & {\f}_{t} \!=\!f(\mathbf{x}_{t-1}) ,  \ \  f(\cdot)  \!\sim\! \mathcal{G} \mathcal{P}\left(\bm{\mu}(\cdot), \bm{k}_{\btheta_{gp}}(\cdot, \cdot)\right), \ \ \mathbf{x}_{0} \!\sim\! p(\mathbf{x}_{0} ), \\
        & \mathbf{x}_{t} \mid {\f}_{t}  \sim \mathcal{N}\left( \x_t \mid {\f}_{t}, \mathbf{Q}\right), \quad  \ \mathbf{y}_{t} \mid  \mathbf{x}_{t} \sim \cN \left(\mathbf{y}_{t} \mid \bm{C} \mathbf{x}_{t}, \boldsymbol{R}\right),
    \end{align}
\end{subequations}
where
the emission model is assumed to be known and linear with the coefficient matrix, $\bm{C}=[\bm{I}_{d_y}, \bm{0}] \in \mathbb{R}^{d_y \times d_x}$, to reduce the system non-identifiability \cite{frigola2015bayesian}. In the case of  $d_x > d_y$, we say that the latent states are \textit{partially observable}. The state transitions and observations are corrupted by zero-mean Gaussian noise with covariance matrices $\bm{Q}$ and $\bm{R}$, respectively.  If the state dimension $d_x>1$, the transition $f(\cdot): \mathbb{R}^{d_x} \rightarrow  \mathbb{R}^{d_x}$ is typically modeled with $d_x$ mutually independent GPs.
More concretely, independent GP priors are placed on each  dimension-specific function $f_d(\cdot): \mathbb{R}^{d_x} \rightarrow \mathbb{R}$, and
\begin{equation}
\setlength{\abovedisplayskip}{3.5pt}
\setlength{\belowdisplayskip}{3.5pt}
\f_t = f(\x_{t-1}) \triangleq \{f_d(\x_{t-1})\}_{d=1}^{d_x} \triangleq \{\f_{t,d}\}_{d=1}^{d_x} , 
\label{eq:multivariate_GP}
\end{equation}
where each independent GP has its own mean function, kernel function, and hyperparameters. The challenging task in GPSSM is to learn the transition function and noise models, i.e., $\btheta=[\btheta_{gp}, \bm{Q}, \bm{R}]$, and infer the latent states of interest simultaneously.  

\vspace{-.05in}
\section{Output-Dependent GPSSM}
\label{sec:proposed-model}
\vspace{-.1in}
In this section, we first point out the issues existing in the GPSSM literature, then propose our output-dependent GPSSM and explain why it is able to overcome these issues. Lastly, we detail the proposed variational learning method for the output-dependent GPSSM. 

\vspace{-.1in}
\subsection{Problem Statement and Output-Dependent GPSSM}
\label{subsec:problem_statement} \vspace{-.05in}

\begin{figure}[t!]
	\centering		
	\subfloat[Independent GPs]{
        \label{fig:MOGP_indep}
        \includegraphics[width =0.185 \textwidth]{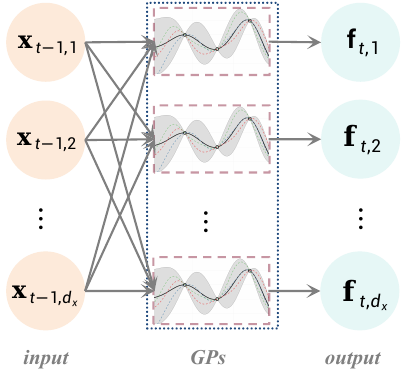}
        } 
        \hfill 
	\subfloat[Dependent GPs using LMC]{
        \label{fig:MOGP}
        \includegraphics[width =0.27\textwidth]{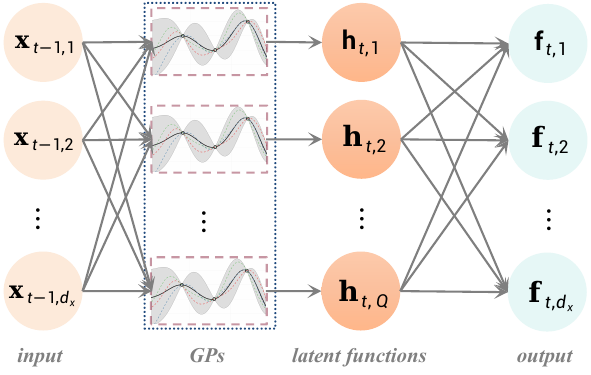}
        }
	\vspace{-.13in}
	\caption{Output-independent GPSSMs vs. output-dependent GPSSMs}
	\vspace{-.2in}
	\label{fig:MOGPSSMs}
\end{figure}

As depicted in Fig.~\ref{fig:MOGP_indep} and described in Section \ref{subsec:GPSSM-review} (see Eq.~(\ref{eq:multivariate_GP})), the existing GPSSM works assume the transition function outputs are independent when modeling high-dimensional latent dynamics.  The adverse effects of independent modeling are twofold. First, model mismatch can occur, especially when there are strong correlations among the outputs. In fact,  high-dimensional latent states in various applications tend to be inherently dependent.  For example, in navigation and tracking applications, the latent states comprise physical quantities such as displacement, acceleration, and velocity \cite{sarkka2013bayesian} that are strongly correlated according to physic law.  In such applications, the independent outputs assumption 
will degenerate the learning and inference performance.  Second, the inductive bias cannot be transferred between outputs, which limits the model learning capacity \cite{caruana1997multitask}, especially when the latent states are only partially observed.


To explicitly model the dependency among outputs of the transition function, we propose to apply the linear model of coregionalization (LMC) \cite{alvarez2012kernels,chen2022multitask}, which is a well-known, simple, yet practical multiple-output GP framework that linearly mixes $Q$ independent latent GPs for modeling multiple dependent outputs simultaneously. More specifically, as depicted in Fig.~\ref{fig:MOGP}, the $d_x$ transition outputs $\{\f_{t, d}\}_{d=1}^{d_x}$ are obtained by the linear combinations of $Q$ independent latent GPs, $\mathbf{h}_t = \{\mathbf{h}_{t, q}\}_{q = 1}^Q$, where $\mathbf{h}_{t, q} = h_q(\x_{t-1}), h_q(\cdot) \sim \mathcal{GP}(0, k_q(\cdot, \cdot))$, i.e., 
\begin{equation}
\setlength{\abovedisplayskip}{2.5pt}
\setlength{\belowdisplayskip}{2.5pt}
  \f_{t,d} \!=\! f_d(\x_{t-1}) \!=\! \sum_{q=1}^{Q} \bm{a}_{d, q} \cdot h_q(\x_{t-1}) \! = \! \bm{a}_{d}^{\top} \mathbf{h}_t,  \  d \!=\! 1,..., d_x,
\label{eq:MOGP_LMC}
\end{equation}
where $\bm{a}_{d} \!=\! [\bm{a}_{d, 1}, \bm{a}_{d, 2}, ..., \bm{a}_{d, Q}]^\top \!\in\! \mathbb{R}^{Q}$ are the dimension-specific coefficients that form the coregionalization matrix $\bm{A} = [\bm{a}_1, ..., \bm{a}_{d_x}]^\top$. 
In this way, the $Q$ latent GPs will learn a shared knowledge (inductive bias) of the underlying dynamics, and $\bm{a}_{d}$ will adapt the behaviours for the dimension-specific output.  
Under the LMC framework and the assumption of $Q$ independent latent GPs, the vector-valued transition function follows a Gaussian process prior $f(\x) \!\sim\! \mathcal{GP}(\bm{\mu}(\x), \bm{k}_{\btheta_{gp}}(\x, \x^\prime))$, where the mean function is $\bm{\mu}(\x) \!=\!\bm{0}$,  
and the multi-output kernel function $\bm{k}_{\btheta_{gp}}(\x, \x^\prime)$ has $d_x^2$ entries, specifically,  $[\bm{k}_{\btheta_{gp}}(\x, \x^\prime)]_{i, j} = \sum_{q = 1}^Q \bm{a}_{i, q} \bm{a}_{j, q} k_q(\x, \x^\prime), i, j = 1,2,..., d_x$. Compared with the classic (independent) GP modeling, 
the GP prior with LMC explicitly models the correlation between different outputs (with extra matrix multiplication operations). 
Thus, the inductive biases among outputs can be transferred to improve the overall model learning and inference capacity, which is beneficial to, e.g., partially observable state inference.
Note that the new set of $\btheta_{gp}$ includes the coregionalization matrix $\bm{A}$ and the hyperparameters from all $Q$ independent latent GPs.  
It is also noteworthy that even though modeling the output dependency improves the GPSSM flexibility,  it also brings model identifiability issues (i.e., given $\f_{t}$ and $\bm{A}$, the underlying $\mathbf{h}_t$ may not be inferred uniquely). An identifiable model is critical to state inference. The following corollary indicates that severe non-identifiability can be eliminated by carefully selecting the $Q$ parameter.
\vspace{-.05in}
\begin{corollary}
\label{corollary1}
The proposed output-dependent GPSSM does not compromise the model identifiability if $Q \le d_x$ and $\operatorname{rank}(\bm{A}) = Q$.
\end{corollary}
\vspace{-.15in}
\begin{proof}
 When $Q \le d_x$ and $\bm{A}$ is full column rank, given $\f_t$ and $\bm{A}$ for the underlying linear system $\bm{A} \mathbf{h}_t = \f_t$, the estimate $\hat{\mathbf{h}}_t = (\bm{A}^\top \bm{A})^{-1} \bm{A}^\top \f_t$ gives an exact solution if true $\mathbf{h}_t$ exists. Then, by following the theorem in Section 3.2.1, \cite{frigola2015bayesian}, severe model idenfiability issue can be solved. 
\end{proof}
\vspace{-.15in}
\begin{remark}
When $Q>d_x$, the output-dependent GPSSM is more flexible, and is potentially beneficial to sequence forecasting (see Section \ref{subsec:realdata}). However, for latent state inference purposes, Corollary \ref{corollary1} suggests setting $Q\le d_x$, which can help avoid the model non-identifiability and improve state inference performance.
\end{remark}
\vspace{-.1in}

\vspace{-.1in}
\subsection{Variational Learning and Inference}
\label{subsec:ODGPSSM} \vspace{-.08in}
{Performing inference for the GPSSM is generally intractable. Instead of using Monte Carlo-based methods, we resort to variational inference methods for computational tractability and efficiency.}
For the convenience of discussion, we denote $\vx = \{\x_{t}\}_{t=0}^{T}$, $\vy = \{\y_{t}\}_{t=1}^{T}$, and $\vf = \{\f_{t}\}_{t=1}^{T}$. 
To alleviate the computation cost of GP models, we use the sparse GP method \cite{hensman2013gaussian}, which introduces a small set of inducing points $\vec{\z} = \{\z_{i}\}_{i=1}^m$ and $\vec{\u} = \{\u_{i, q}\}_{i, q=1}^{m, Q}$ to serve as the surrogate of the associated GPs, where $\u_{i, q} = h_{q}(\z_{i}) \in \mathbb{R}$ 
and $\z_i \in \mathbb{R}^{d_x}$. Here, we assume the $Q$ inducing outputs $\{\u_{i,q}\}_{ q=1}^Q$ share the inducing inputs $\z_i$, since all the latent GPs take the shared input $\x_{t-1}$ at any time step $t$, as depicted in Fig.~\ref{fig:MOGP}. Sharing inducing inputs between different latent GPs will also reduce the number of variational parameters. 
Based on these settings, the joint distribution of the output-dependent GPSSM augmented with inducing points is 
\begin{equation}
\setlength{\abovedisplayskip}{3pt}
\setlength{\belowdisplayskip}{3pt}
\begin{aligned}
\!\! p(\vy, \vx, \vf, \vu) \!=\! p(\x_{0}) \prod_{t =1}^{T}  p(\y_{t} \vert \x_{t}) p(\x_{t} \vert \f_{ t})p(\f_{t} \vert \x_{t-1}, \vu) p(\vu),
\end{aligned}
\end{equation}
where $p(\vu) = \prod_{q = 1}^Q p(\{\u_{i, q}\}_{i=1}^m)$ due to the independence assumption. The distribution of the transition function outputs $p(\f_{t} \vert \x_{t-1}, \vu)$ are determined by the latent GPs $p(\h_t \vert \x_{t-1}, \vu)$, where
\begin{equation}
\setlength{\abovedisplayskip}{3.5pt}
\setlength{\belowdisplayskip}{3.5pt}
p(\mathbf{h}_t \vert \x_{t-1}, \vu) \!=\! \prod_{q = 1}^Q \cN(\mathbf{h}_{t,q} \vert \bm{\xi}_{\h_{t,q}}, \bm{\Xi}_{\h_{t,q}}), 
\end{equation}
and $\cN(\bm{\xi}_{\h_{t,q}}, \bm{\Xi}_{\h_{t,q}})$ is the $q$-th GP posterior distribution with $\x_{t-1}$ as test input while $(\vz, \ \{\u_{i, q}\}_{i=1}^m)$ as training data, see Eqs.~(\ref{eq:posterior_mean}) and (\ref{eq:posterior_vairiance}). 
Therefore, 
\begin{equation}
\setlength{\abovedisplayskip}{3pt}
\setlength{\belowdisplayskip}{3pt}
    p(\f_{t} \vert \x_{t-1}, \vu) \!=\! \int_{\h_t} p(\f_t \vert \h_t) \ p(\h_t \vert \x_{t-1}, \vu) \!=\!  \cN(\f_t \vert \bm{\xi}_{\f_t}, \bm{\Xi}_{\f_t}),
    \label{eq:transition_prior}
\end{equation}
where $\bm{\xi}_{\f_t}  = \bm{A} \bm{\xi}_{\h_{t}},  \ \bm{\Xi}_{\f_t}  = \bm{A} \bm{\Xi}_{\h_{t}} \bm{A}^\top$ and $\bm{\xi}_{\h_{t}} = [\bm{\xi}_{\h_{t,1}}, .., \bm{\xi}_{\h_{t,Q}}]^\top$, $\bm{\Xi}_{\h_{t}}  = \operatorname{diag}(\bm{\Xi}_{\h_{t,1}}, ..., \bm{\Xi}_{\h_{t,Q}})$.

The main idea behind the variational inference method is to approximate the intractable posterior distribution $p(\vec{\x}, \vec{\f}, \vec{\u} \vert \vy)$ using a variational distribution $q(\vec{\x}, \vec{\f}, \vec{\u})$,  leading to the evidence lower bound (ELBO),  $\mathcal{L} \triangleq \mathbb{E}_{q(\vec{\x}, \vec{\f}, \vec{\u})}\left[  \log \frac{p(\vy, \vx, \vf, \vu)}{q(\vec{\x}, \vec{\f}, \vec{\u})}\right] \le \log p(\vy)$.  
Different choices of  the variational distribution induce different ELBOs, hence different learning algorithms for GPSSM \cite{frigola2015bayesian}. In this paper, we choose the variational distribution $q(\vec{\x}, \vec{\f}, \vec{\u})$ in the form of 
$
    q(\x_{0}) \prod_{t =1}^{T} p(\x_{t} \vert \f_{t}) p(\f_{t} \vert \x_{t-1}, \vu) q(\vu),
$
where 
\begin{equation}
\setlength{\abovedisplayskip}{2.5pt}
\setlength{\belowdisplayskip}{2.5pt}
    q(\vu) 
    = \prod_{q=1}^Q \cN(\{\u_{i, q}\}_{i=1}^m \vert \m_q, \mathbf{S}_{q}) 
    = \cN(\vu \mid \vec{\m}, \mathbf{S}),
    \label{eq:qu_variational}
\end{equation}
and moreover the mean vector 
$\vec{\m} = [\m_1^\top, ..., \m_Q^\top]^\top \in \mathbb{R}^{m Q}$, and the covariance matrix $\mathbf{S} = \operatorname{diag}(\mathbf{S}_1, ..., \mathbf{S}_Q)\in \mathbb{R}^{m Q \times m Q}$,
are free variational parameters. The variational distribution for the initial state is parameterized by a recognition network with input, $\vy$, and parameters, $\bzeta$, i.e.,  $q_{\bzeta}(\x_{0}) = \mathcal{N}(\x_0 \vert \m_{\x_{0}}, \bm{S}_{\x_{0}})$, where $\m_{\x_{0}}, \bm{S}_{\x_{0}}$ are the outputs of the recognition network \cite{eleftheriadis2017identification}.
Note that the form of variational distribution selected in this paper is similar to the one used in the probabilistic SSM  \cite{doerr2018probabilistic}. However, instead of assuming independent outputs for the transition function, the GPSSM prior as well as the approximated posterior considered in this paper explicitly construct the dependency among outputs by mixing $Q$ latent GPs, thus making the proposed model more realistic and flexible. After some algebraic calculations, the corresponding ELBO becomes
\begin{equation}
\setlength{\abovedisplayskip}{2.5pt}
\setlength{\belowdisplayskip}{2.5pt}
   \begin{aligned}
    \mathcal{L}(\btheta) = \sum_{t=1}^{T} \ & \mathbb{E}_{q(\x_{t})} \left[ \log p(\y_{t} \vert \x_{t})\right] - \operatorname{KL}\left[ q(\vu) \| p(\vu)\right] \\
    & - \operatorname{KL}\left[ q(\x_{0}) \| p(\x_{0})\right],
   \end{aligned}
   \label{eq:ELBO_final}
\end{equation}
where the first term encourages decent samples drawn from the variational distribution $q(\x_{t})$ to fit the emission model well, while the second and third terms regularize the initial state, and the posterior distributions of the $Q$ latent GPs thus the posterior of $f(\cdot)$, respectively.
The two latter terms can be computed analytically, however, the expectation terms, $\mathbb{E}_{q(\x_{t})} \left[ \log p(\y_{t} \vert \x_{t})\right], \forall t,$ 
need to be evaluated by sampling method and reparametrization trick \cite{kingma2019introduction} due to the intractability of $q(\x_{t})$ \cite{doerr2018probabilistic}. 
The sampling steps are described as follows.
By conditioning on the latent state $\x_{t-1}$, we have
\begin{equation}
\setlength{\abovedisplayskip}{3.5pt}
\setlength{\belowdisplayskip}{3.5pt}
\begin{aligned}
    q(\x_{t} \vert \x_{t-1}) & = \int_{\f_{t}, \vu}  p(\x_{t} \vert \f_{ t})p(\f_{t} \vert \x_{t-1}, \vu) q(\vu)\\
    & =\cN(\x_{t} \mid  {\m}_{t|t-1}, \ \mathbf{S}_{t|t-1} ),
\end{aligned}
\label{eq:state_conditional}
\end{equation}
where ${\m}_{t|t-1} = \bm{A} \m_{\h_t}$, $\mathbf{S}_{t|t-1} = \bm{A} \mathbf{S}_{\h_t} \bm{A}^\top + \bm{Q}$, and  $\m_{\h_t} = [\m_{\h_{t,1}}, ..., \m_{\h_{t,Q}}]^\top, \mathbf{S}_{\h_t} = \operatorname{diag}(\mathbf{S}_{\h_{t,1}}, ..., \mathbf{S}_{\h_{t,Q}})$ with 
$$
\setlength{\abovedisplayskip}{3.5pt}
\setlength{\belowdisplayskip}{3.5pt}
\left\{
    \begin{aligned}
        & \m_{\h_{t,q}} = \k_{\x_{t-1}, \vz} \ \k_{\vz, \vz}^{-1} \ {\m_q},  \\
        & \mathbf{S}_{\h_{t,q}} \!=\! \k_{\x_{t-1},\x_{t\!-\!1}} \!-\! \k_{\x_{t\!-\!1}, \vz} \ \k_{\vz, \vz}^{-1}\left[\k_{\vz, \vz} - \mathbf{S}_q \right]\k_{\vz, \vz}^{-1} \ \k_{\x_{t\!-\!1}, \vz}^\top, 
    \end{aligned}
\right.
$$
for any $q$. Note that here we omit the subscript $q$ in the kernel matrix of the $q$-th latent GP for notation brevity.
Using Eq.~(\ref{eq:state_conditional}) we can recursively sample latent states $\x_t, t = 1,2,..., T$, by starting from sampling $\x_0 \sim q(\x_0)$, so that the ELBO can be numerically evaluated. Together with all, we apply stochastic gradient ascent and use the Adam optimizer to maximize the lower bound $\mathcal{L}(\btheta)$ over parameters 
$\btheta = [\bzeta, \vz, \vec{\m}, \mathbf{S}, \btheta_{gp}, \bm{Q}, \bm{R}]$.  The gradient can be propagated back through time owing to the chain rule sampling and reparametrization trick \cite{kingma2019introduction}, and the parameters will converge to a stationary point.
\vspace{-.1in}
\begin{remark}[Computational Complexity] 
\label{remark:comput_complexity}
Typically, the number of GP inducing points, $m$, is larger than the number of latent GPs, $Q$, and the state dimension, $d_x$. For a data sequence with the length of \ $T$ and assuming $T \!  \gg  \! m  \! >\! Q  \! \ge \! d_x$,  the computational complexity of  evaluating the ELBO (Eq.~(\ref{eq:ELBO_final})) scales as 
 $\mathcal{O}{(T Q m^2 \!+\!  {T  Q^2  d_x}})$.
Compared with the output-independent GPSSM \cite{doerr2018probabilistic} that scales as $\mathcal{O}{(T d_x m^2)}$,
  we can observe that only gentle computational complexity increases in the output-dependent GPSSM (especially in the case of $Q \! = \! d_x$).
\end{remark} \vspace{-.12in}
%
%

\vspace{-.05in}
\section{Experimental Results}
\label{sec:experimental-results} 
\vspace{-.1in}
In this section, we show the performance of the proposed \textit{output-dependent} GPSSM (termed \textit{ODGPSSM}) on one synthetic dataset and five real system identification datasets. For comparison, we choose two \textit{output-independent} baseline models: 1)  probabilistic recurrent SSM (\textit{PRSSM}) \cite{doerr2018probabilistic}; and 2)  deep state-space model (\textit{DSSM}) \cite{krishnan2017structured}. 

\vspace{-.1in}
\subsection{Synthetic Dataset}
\label{subsec:syntheticdata} \vspace{-.05in}
\begin{figure}[t!]
    \centering		
    \subfloat[The latent space of PRSSM ]
    {
    \label{fig:MOGPSSM_indep}
    \includegraphics[width =.48\textwidth]{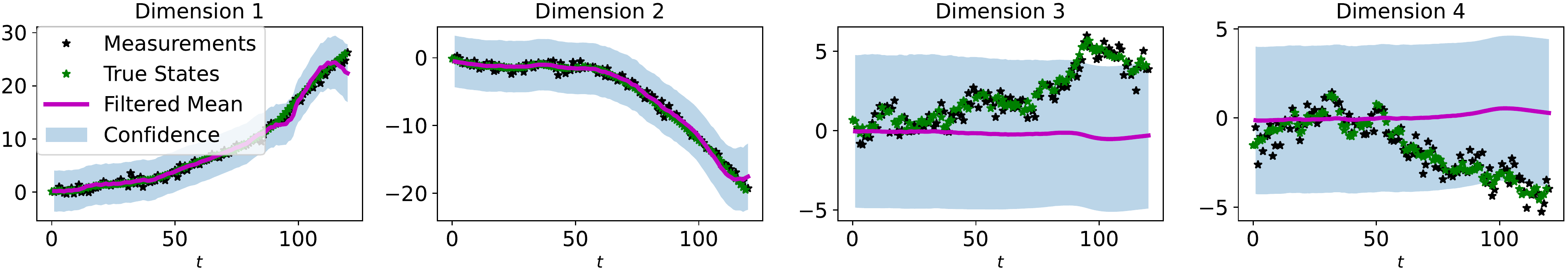}
    } 
    \vspace{-.1in} 
    
    \subfloat[ The latent space of ODGPSSM ]
    {
    \label{fig:MOGPSSM_LMC}
    \includegraphics[width =.48\textwidth]{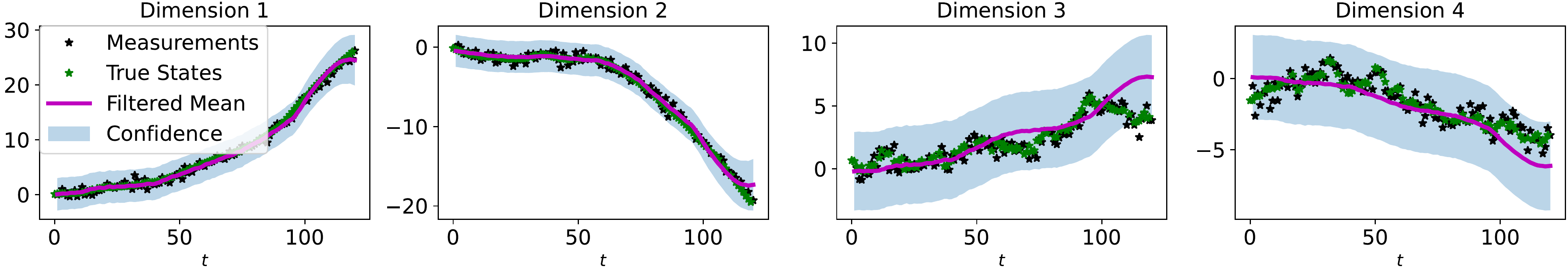}
    }
    \vspace{-.09in}
    \caption{The latent space learned by two different GPSSMs}
    \vspace{-.1in}
    \label{fig:MOGPSSM_results}
\end{figure}
We first use a simple and stylized numerical 2-dimensional (2-D) car tracking example provided in \cite{sarkka2013bayesian} (Example 4.3) to show the importance of modeling dependency between the transition outputs. More concretely, in this example, the underlying car dynamic is characterized by the linear Gaussian state-space model (LGSSM),
$$
\setlength{\abovedisplayskip}{4.5pt}
\setlength{\belowdisplayskip}{4.5pt}
\begin{aligned}
    & \x_{t} = \begin{bmatrix}
        \bm{I}_2  &\bm{I}_2 \\
        \bm{0}    & \bm{I}_2 \\
    \end{bmatrix} \x_{t-1} + \mathbf{v}_t,  \quad \mathbf{v}_t\sim \cN(\bm{0}, \bm{Q}),\\
    & \y_t = [\bm{I}_{2}, \bm{0}] \  \x_{t} + \mathbf{e}_t, \qquad \quad  \mathbf{e}_t\sim \cN(\bm{0}, \bm{R}),
\end{aligned}
$$
where the {partially observable} state $\x_{t} \in \mathbb{R}^{4}$ consists of 2-D car positions and 2-D velocities; $\y_t$ is a noisy observation of the car positions. 
For more details about this model, one can refer to \cite{sarkka2013bayesian}.  
Note that the entries of the state vector, $\{\x_{t,d}\}_{d=1}^4$, are correlated due to the linearity and Gaussianity of the state transition in the LGSSM. {Hence, by exploiting the correlations with the observed states $\{\x_{t,d}\}_{d=1}^2$, it is possible to infer the unobserved ones, $\{\x_{t,d}\}_{d=3}^4$.}

We use the underlying LGSSM to generate $T$=120 observations $\vy =\{\y\}_{t=1}^T$ for training the \textit{PRSSM} \cite{doerr2018probabilistic} and the newly proposed \textit{ODGPSSM}. For both PRSSM and ODGPSSM, we adopt the following initialization: 1) The initial state, $\x_{0}\!=\! [0,0, 1, -1]^\top$, is assumed to be known, hence there is no need to train the recognition network; 2) The dimension of the latent state, $d_x$, is set to be $4$, and the number of the GP inducing points $m$ is set to be $20$; 3) The GP transition models are pretrained/initialized using $20$ true latent state pairs with the same training epochs.  The number of the latent GP functions for ODGPSSM remains the same as the state dimension, 
thus only slightly increasing the overall  computational complexity (see Remark \ref{remark:comput_complexity}). Fig.~\ref{fig:MOGPSSM_results} depicts the learning results of ODGPSSM and PRSSM. It can be observed that both PRSSM and ODGPSSM infer the first two dimensions of the latent states well. However, the PRSSM fails to capture the underlying dynamics in the 3rd and 4th unobserved dimensions.  {This is due to the fact that the independent modeling in PRSSM ignores the correlations between the states, resulting in the fluctuations of the GP transition posterior around the zero-mean prior.
In contrast, ODGPSSM establishes the dependencies through the LMC framework, making it capable of correctly learning the GP transition posterior and inferring the unobserved states by exploiting the knowledge from the shared correlations and the first two dimensions that are fully observed.}

%

\vspace{-.1in}
\subsection{Real Datasets}
\label{subsec:realdata} 
\vspace{-.05in}
Since the superiority of PRSSM compared to classic time-series modeling approaches has been shown in \cite{doerr2018probabilistic}, we will skip similar comparisons due to space limitations. In this subsection, we only compare ODGPSSM with its two competitors, PRSSM and DSSM, on five real system identification datasets introduced in \cite{doerr2018probabilistic} (see \cite{doerr2018probabilistic} for more details). For each dataset, the first half of a sequence is used for training and the rest for testing.  All datasets are standardized using training sequence and the latent state dimension is set to be $d_x = 4$.
More detailed model settings can be found in the accompanying code online available at: \textit{https://github.com/zhidilin/ODGPSSM}.
For DSSM, we refined the code from a {public implementation} at \textit{http://pyro.ai/examples/dmm.html}, and set the emission function to be the same as the two GPSSMs. The test results are reported in Table \ref{tab:mse-SIdate}, where the root-mean-square error (RMSE) is averaged over 100-step ahead forecasting.
%
\begin{figure}[t!]
    \centering     \includegraphics[height=3.5cm, width=0.4\textwidth]{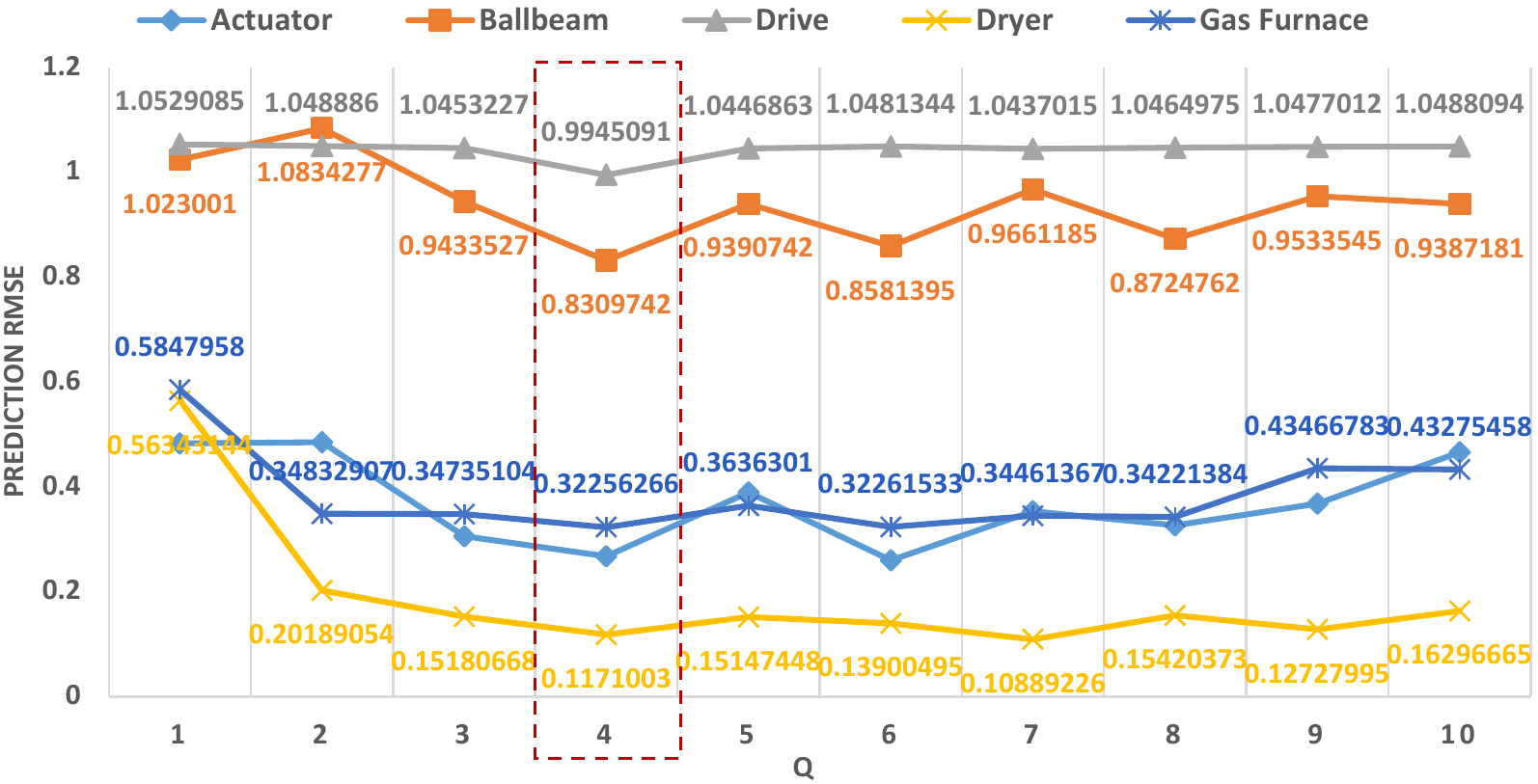}
    \vspace{-.1in}
    \caption{Test performance of ODGPSSM under different values of $Q$}
    \vspace{-.07in}
    \label{fig:my_label}
\end{figure}
\begin{table}[t!]
    \begin{center}
        \caption{Prediction RMSE comparison (on standardized test sets) between the proposed ODGPSSM (with $Q=4$) and its competitors.}
        \vspace{-.1in}
        \label{tab:mse-SIdate}
        \begin{adjustbox}{width=0.48\textwidth}
        \begin{tabular}{cccccc}
            \toprule
            Methods & 
            \begin{tabular}[c]{@{}c@{}} Actuator \end{tabular} & 
            \begin{tabular}[c]{@{}c@{}} Ballbeam \end{tabular} &
            \begin{tabular}[c]{@{}c@{}} Drive \end{tabular} &
            \begin{tabular}[c]{@{}c@{}} Dryer \end{tabular} & 
            \begin{tabular}[c]{@{}c@{}} Gas Furnace \end{tabular} \\
            \midrule
            DSSM  
            &  3.0752
            &  2.6311
            &  1.8417
            &  2.7055
            &  1.7746\\ 
            PRSSM
            &  0.3592  
            &  0.9082 
            &  1.0459  
            &  0.1334  
            &  0.3356    \\ 
            ODGPSSM  
            &  \textbf{0.2668}  
            &  \textbf{0.8309}  
            &  \textbf{0.9945}  
            &  \textbf{0.1171}  
            &  \textbf{0.3225} \\
            \bottomrule
        \end{tabular}
        \end{adjustbox}
    \end{center}
    \vspace{-.3in}
\end{table}

From Table \ref{tab:mse-SIdate} we can observe that ODGPSSM consistently outperforms PRSSM in terms of the prediction RMSE across all the datasets, which convincingly illustrates the benefits of output dependency modeling.  It can also be observed that both ODGPSSM and PRSSM outperform DSSM in terms of the prediction RMSE.
The reason is that both the transition function and the variational distributions in DSSM are modeled by deep neural networks that require big data to tune the large number of parameters. However, in our case, the datasets are relatively small, e.g. the training set of the \textit{Gas Furnance} dataset is merely of length 148,  which is insufficient to support the DSSM learning. In contrast, GPSSMs inherit the merits of GP, showing unique superiority in small dataset regimes. 

Finally, we investigate the impact of $Q$ in ODGPSSM, since the number of latent GPs, $Q$, determines the model flexibility.
The results in Fig.~\ref{fig:my_label} show that the prediction RMSE across almost all the datasets reach the lowest points when $Q \!\!=\!\! d_x \!\!=\! \! 4$, {even though the models with $Q\!\!>\!\!4$ is more flexible than the ones with $Q\!=\!4$}. This is probably because the additional parameters (additional coregionalization coefficients and variational parameters)
makes the model unidentifiable and the learning more difficult.  Future work will attempt to remedy this problem by introducing sparse constraints on the coregionalization coefficients and verify it on real data provided by China Unicom.

\vspace{-.1in}
\section{Conclusion}
\label{sec:conclusion} 
\vspace{-.1in}
In this paper, we propose an output-dependent GPSSM by explicitly modeling the output dependency of GP transition using the well-known, simple yet practical LMC framework. We also propose a variational learning algorithm that only gently increases the computational complexity to learn the output-dependent GPSSM. Experimental results show that modeling the output dependency in GPSSM not only facilitates latent state inference when the latent state is partially observed, but also makes the GPSSM more competent than its competitors in terms of prediction.

\bibliographystyle{IEEEbib}
\bibliography{refs}

\end{document}